\documentclass[conference]{IEEEtran}

\usepackage{times}
\usepackage{latexsym}
\usepackage[T1]{fontenc}
\usepackage[utf8]{inputenc}
\usepackage{microtype}
\usepackage{inconsolata}
\usepackage{graphicx}
\usepackage{booktabs}
\usepackage{multirow}
\usepackage{array}
\usepackage{placeins}
\usepackage{cite}
\usepackage{url}
\usepackage{xcolor}

\usepackage{subcaption}
\usepackage[hidelinks]{hyperref}

\graphicspath{{Figures/}}

\title{Beyond Pass Rate: A Multilingual, Execution-Grounded Evaluation of Open Code LLMs}

\author{
\IEEEauthorblockN{Sayed Erfan Arefin}
\IEEEauthorblockA{
Department of Computer Science\\
University of Dayton\\
Dayton, OH 45469, USA\\
Email: sarefin1@udayton.edu
}
}

\begin{document}

\maketitle

\begin{abstract}
Code generation models are typically compared using compact execution benchmarks and aggregate pass rates, but such summaries obscure how performance varies across programming languages, problem families, and failure modes. We present a large-scale, execution-grounded evaluation of 9 openly accessible LLMs specialized for coding on 2,707 free LeetCode problems across 12 programming languages. Our corpus contains 325,343 problem--model--language jobs, each linked to prompt metadata, extracted code, LeetCode execution outcomes, and static-analysis signals. The results show that current open models remain far from the human acceptance reference: the best model, Yi-Coder-9B-Chat, reaches 23.64\% mean correctness, compared with a 57.2\% human acceptance baseline. Rankings are also slice-dependent: Qwen2.5-Coder-14B-Instruct is strongest on hard problems and distinct-problem coverage, while Gemma-2-27B-IT achieves the highest all-language lint pass rate. Failure analysis shows that compile errors account for 63.25\% of non-accepted best submissions, indicating that many failures occur before semantic correctness can be tested. Static quality further diverges from functional correctness. Together, these findings show that multilingual, artifact-preserving evaluation reveals tradeoffs hidden by single-language or single-metric leaderboards.
\end{abstract}

\begin{IEEEkeywords}
code generation, large language models, execution-grounded evaluation, multilingual benchmarks, LeetCode
\end{IEEEkeywords}

\section{Introduction}
\label{sec:introduction}

Large language models (LLMs) for code are increasingly used as programming assistants, but standard evaluations often rely on compact function-level benchmarks, single-language settings, or aggregate pass@1-style scores. These evaluations are useful, yet they obscure properties that matter in practice: whether a model is reliable across languages, how performance changes with task difficulty, whether failures are syntactic or semantic, and whether accepted code is also maintainable. We study these questions using a large execution-grounded archive of LeetCode-derived tasks. The corpus contains 2,707 free LeetCode problems, 325,343 cleaned problem--model--language jobs across 12 programming languages and 9 open or openly accessible code LLMs. Each job preserves the prompt, target language, model identity, raw response, extracted code, official LeetCode result, and static-analysis signals, enabling analysis beyond binary accepted/not-accepted outcomes. Our contributions are three fold:
\begin{itemize}
    \item We construct and analyze a large scale execution archive covering 2,707 problems, 12 programming languages and 9 models.
    \item We evaluate open code LLMs across mean correctness, accepted-problem coverage, difficulty, language, topic, pairwise head-to-head outcomes, and failure composition.
    \item We jointly analyze functional correctness and static quality, showing that execution success and maintainability-oriented signals can diverge substantially.
\end{itemize}

Our results show that model comparison is multidimensional. Yi-Coder-9B-Chat achieves the highest mean correctness and strongest average language rank, while Qwen2.5-Coder-14B-Instruct performs best on hard problems and accepted-problem coverage. All models remain far below the human acceptance reference, compile errors dominate non-accepted outcomes, and static quality does not follow the same ordering as functional correctness. These findings motivate code-model evaluation that preserves execution, language coverage, failure modes, and quality signals instead of collapsing performance into a single score.

\section{Related Research}
\label{sec:related}

Execution-based evaluation remains a central method for studying code generation because it checks whether generated programs actually run and satisfy task-level tests. Chen et al.~\cite{chen2021codex} evaluated large language models trained on code using executable programming tasks, while Austin et al.~\cite{austin2021mbpp} studied program synthesis with large language models and Hendrycks et al.~\cite{hendrycks2021apps} measured coding-challenge competence using APPS-style programming problems. Li et al.~\cite{li2022alphacode,li2022competition} extended this direction to competition-level code generation, and Lu et al.~\cite{lu2021codexglue} introduced CodeXGLUE as a multi-task benchmark suite for code understanding and generation. CodeBLEU prposed by Ren et al.~\cite{ren2020codebleu} as an automatic code-evaluation metric that incorporates lexical, syntactic, and semantic signals, but execution-based judging remains important because it directly measures functional behavior.

Subsequent work broadened code-model evaluation across languages, task types, and freshness settings. Cassano et al.~\cite{cassano2022multiple}, Zheng et al.~\cite{zheng2023codegeex}, Athiwaratkun et al.~\cite{athiwaratkun2022mbxp}, and Jain et al.~\cite{jain2024livecodebench} studied multilingual or contamination-aware evaluation settings, showing that code-model performance should not be treated as a single-language or single-benchmark property. Lai et al.~\cite{lai2022ds1000}, Du et al.~\cite{du2023classeval}, Gu et al.~\cite{gu2024cruxeval}, and Zhuo et al.~\cite{zhuo2024bigcodebench} further expanded evaluation toward data-science code, class-level generation, code reasoning, pragmatic code generation, and complex function-call tasks. Puri et al.~\cite{puri2021codenet} introduced a large-scale code dataset covering diverse programming tasks, while Roziere et al.~\cite{roziere2020transcoder} and Lachaux et al.~\cite{roziere2020transcoder} studied unsupervised translation between programming languages. Together, these studies motivate evaluating code models across languages, problem categories, and execution outcomes rather than relying on one narrow benchmark.

Repository-level and context-rich benchmarks form another related thread. Liu et al.~\cite{liu2023repobench} studied repository-level code completion, Ding et al.~\cite{ding2023crosscodeeval} focused on cross-file and multilingual code-completion evaluation, Jimenez et al.~\cite{jimenez2023swebench} evaluated whether language models can resolve real GitHub issues, and Zan et al.~\cite{zan2024swebenchjava,zan2025multiswebench} extended issue-resolution evaluation toward Java and multilingual settings. These studies are complementary to competitive-programming evaluations: repository benchmarks test tool use, retrieval, and patch integration, whereas LeetCode-style tasks stress algorithmic reasoning, language-specific syntax, and executable correctness under standardized problem specifications.

Open model development evolved alongside these benchmarks. Feng et al.~\cite{feng2020codebert}, Guo et al.~\cite{guo2021graphcodebert} and Wang et al.~\cite{wang2021codet5} developed representation-learning or encoder-decoder models for program understanding and generation.Subsequent larger generative systems further advanced open code generation. Nijkamp et al.~\cite{nijkamp2022codegen}, Fried et al.~\cite{fried2022incoder}, Allal et al.~\cite{allal2023santacoder}, Luo et al.~\cite{luo2023wizardcoder}, Roziere et al.~\cite{roziere2023codellama}, and Guo et al.~\cite{guo2024deepseekcoder} introduced open or openly described code-generation models with increasingly strong multilingual and instruction-following capabilities. Work on evaluation rigor also showed that benchmark conclusions can be fragile. Liu et al.~\cite{liu2023evalplus} showed that weak test suites can materially overestimate correctness.

Another line of work focuses on \emph{how} code models fail rather than only how often they pass. Tian et al.~\cite{leetcode2}, Arefin et al.~\cite{arefin2024unmasking}, Heya et al.~\cite{ieee2024stable}, Sobania et al.~\cite{bug1}, and Wang et al.~\cite{wang2024deprecatedapis} examined issues such as wrong answers, runtime failures, debugging behavior, repair loops, hallucinations, API/tool use, repeated-query stability, and deprecated API usage. Chen et al.~\cite{chen2022codet}, Shinn et al.~\cite{shinn2023reflexion}, and Yao et al.~\cite{yao2023react} studied test generation, feedback, and reasoning-action loops as mechanisms for improving model outputs. These methods motivate evaluating not only final pass/fail outcomes but also the failure states that would inform repair, debugging, or agentic follow-up systems.

Code quality is a smaller but growing literature. Tian et al.~\cite{leetcode2} used Pylint-style analysis when studying ChatGPT-generated Python solutions, while Simoes and Venson~\cite{simoes2024codequality} compared large language models as code-quality judges against static-analysis tools. However, most existing code-generation benchmarks emphasize functional correctness, and most static-quality studies focus on narrower language slices or smaller model sets. This leaves a gap between execution-oriented benchmarking and maintainability-oriented static analysis.

This study is unique in combining these lines at scale. Rather than evaluating a small function-level benchmark, a single language, a single proprietary model, we analyze a large archive of multiple open coding-LLM submissions across many LeetCode problems, and multiple programming languages. The study jointly preserves functional execution outcomes, difficulty/topic/language slices, accepted-problem coverage, non-accepted failure composition, partial-correctness signals, and static-quality/linter evidence. This makes the work distinct from prior benchmarks, where it does not only ask which model solves more tasks, but also where each model succeeds, which languages and topics change the ranking, how failures manifest, and whether executable success aligns with maintainable code quality.

\section{Methodology}
\label{sec:methodology}

\subsection{Problem Collection and Corpus Construction}

We constructed the benchmark from publicly available, free LeetCode problems. The collection process first enumerated all non-premium problems available through LeetCode and stored the problem statements, metadata, slugs, difficulty labels, topic annotations, and acceptance-rate fields as local text and structured records. We used only free problems so that the resulting benchmark would be reproducible without relying on paid problem access. Each problem statement was preserved before model prompting so that all models received the same task specification for a given problem--language pair.

The benchmark targets 12 programming languages: C, C++, C\#, Go, Java, JavaScript, PHP, Python, Ruby, Scala, Swift, and TypeScript. For each problem, we generated language-specific prompts that instructed the model to produce a LeetCode-compatible solution in the target language. These prompts were stored in a MySQL database as individual execution jobs. Each job MYSQL entry stored one \emph{problem--language--model} attempt and contains the prompt text, problem identifier, language identifier, model identifier, job status, and execution metadata. The database therefore acted as the central source of data for the experiments, supporting reproducible scheduling, retries in case of any operational errors, and later analysis.

\subsection{Distributed Prompt Serving and Model Inference}

To execute the large number of problem--language--model jobs, we implemented a server-backed job queue. The server exposed pending prompts to client workers and tracked each job through a finite set of operational states, including pending, completed, and operational-error states. Client workers repeatedly requested available jobs, submitted the prompt to the selected local model through LM Studio, stored the raw model response, and returned completion metadata to the server. Table~\ref{tab:evaluated-models} summarizes the nine evaluated open-weight or openly accessible instruction-tuned models. We used the default setting of each models in our experiments. The machine that was used in these experiments is a Lambda Vector One machine with a AMD Ryzen 7 CPU, 128 GB Memoery, 4 TB of SSD and Nvidia RTX 4090 GPU with 24 GB of memoery.

\begin{table*}[t]
\centering

\captionsetup{width=0.8\textwidth}
\caption{Evaluated models and high-level model characteristics. The descriptions are used only to identify the model family and intended emphasis; model performance is measured empirically in our benchmark rather than inferred from these descriptions.}
\label{tab:evaluated-models}
\resizebox{0.9\textwidth}{!}{%
\begin{tabular}{lll}
\toprule
\textbf{Model} & \textbf{Producer} & \textbf{Notable characteristic} \\
\midrule
Codestral-22B-v0.1 & Mistral AI & Code-specialized 22B-parameter model \\
DeepSeek-Coder-V2-Lite-Instruct & DeepSeek-AI & Lite variant from the DeepSeek-Coder-V2 family.\\
Gemma-2-27B-IT & Google & 27B-parameter member of the Gemma 2 family. \\
Granite-3.1-8B-Instruct & IBM & 8B-parameter model from IBM's Granite family. \\
Meta-Llama-3.1-8B-Instruct & Meta & 8B-parameter Llama 3.1 model. \\
Phi-4 & Microsoft & 14B-parameter small language model. \\
Qwen2.5-Coder-14B-Instruct & Alibaba/Qwen & 14B-parameter instruction model. \\
Stable-Code-Instruct-3B & Stability AI & Small 3B-parameter model for code-completion. \\
Yi-Coder-9B-Chat & 01.AI & 9B-parameter code model. \\
\bottomrule
\end{tabular}}
\end{table*}

Operational failures were handled separately from model failures. Network interruptions, local inference failures, API timeouts, database errors, or browser automation errors were marked as operational errors rather than treated as incorrect model outputs. After each execution round, jobs in operational-error states were reset and re-queued in later rounds. This retry policy was designed to prevent infrastructure instability from being conflated with model quality. Completed jobs retained the original prompt, raw model response, extracted code artifact, model identifier, language identifier, and timestamps needed for later auditing.

\subsection{Code Extraction and LeetCode Submission}

Model outputs often contained explanations, markdown formatting, or surrounding commentary in addition to code. We therefore applied a post-processing step to extract the code block from each response, specifically the portion enclosed in triple quotes, and saved it as the candidate submission for the corresponding problem and language. The raw response was retained for traceability. Candidate solutions were then evaluated using LeetCode's official execution environment. A Python automation script used Selenium with a Chromium command-line browser session to load each extracted solution and submit it to LeetCode. For each submission, the script recorded the returned submission ID and queried LeetCode's GraphQL API to retrieve the official judge result, including labels such as Accepted, Wrong Answer, Compile Error, Runtime Error, Time Limit Exceeded, and Memory Limit Exceeded. This ensured that correctness was measured by LeetCode's own judge rather than a local approximation of its test suite. Table~\ref{tab:corpus-summary} summarizes the final analysis set.


\begin{table}[t]
\centering
\captionsetup{width=\columnwidth}

\caption{Corpus summary after cleanup.}
\label{tab:corpus-summary}
\begin{tabular}{lr}
\toprule
Statistic & Count \\
\midrule
Problems & 2,707 \\
Models & 9 \\
Programming languages & 12 \\
Total programming jobs & 325,343 \\
Accepted best submissions & 38,761 \\
\bottomrule
\end{tabular}
\end{table}

Because jobs could be retried after operational resets, we avoid double-counting by selecting one best submission per job: highest correctness ratio, then accepted status, then latest timestamp. Correctness is 1.0 for accepted submissions, the passed/total testcase ratio when available, and 0.0 otherwise. We report five metric families: corpus-side metrics for language, difficulty, topic, and acceptance-rate coverage; functional metrics for correctness, accepted coverage, language, difficulty, topic, and language rank; failure metrics for status and wrong-answer severity; comparative metrics for head-to-head outcomes and shared-task significance tests; and static-quality metrics for complexity, function length, accepted-code linting, and all-language \texttt{ast-grep} pass rates. LeetCode acceptance rate is used only as a problem-level human-reference signal. Runtime and memory fields are present but too inconsistently preserved for a defensible efficiency comparison.

\section{Experimental Results}
\label{sec:analysis}

\subsection{Problem Corpus Characteristics}
\label{sec:problem-analysis}

\begin{figure*}[!tbp]
\centering

\begin{subfigure}[t]{0.58\linewidth}
    \centering
    \includegraphics[width=\linewidth]{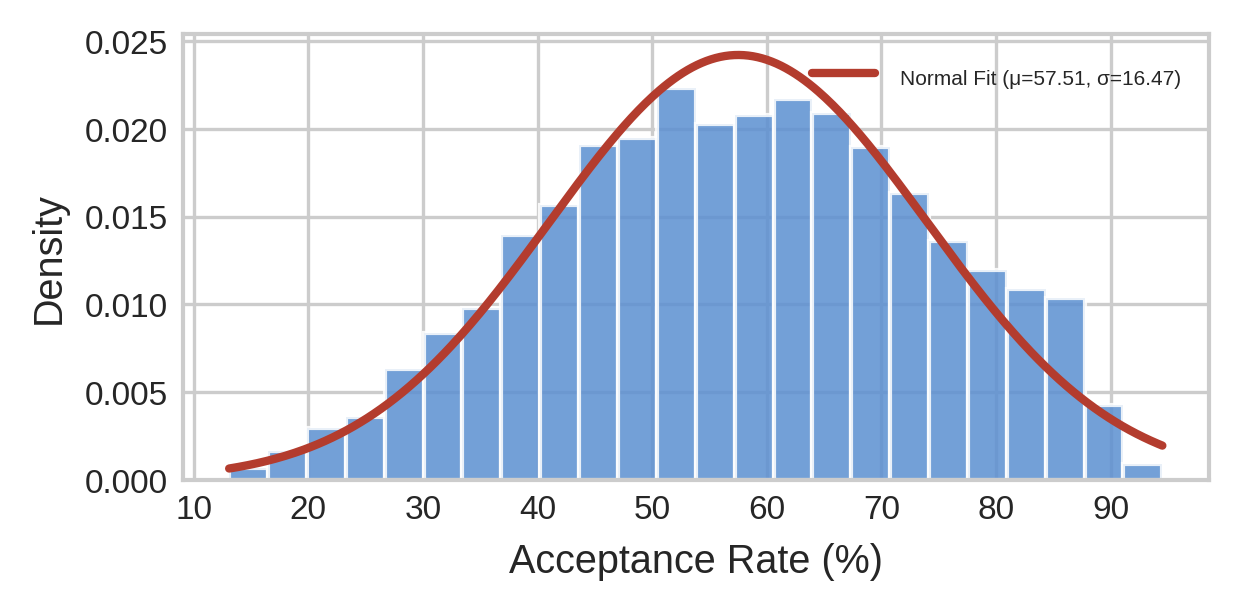}
    \caption{Human acceptance-rate distribution across benchmark problems, summarized by a fitted normal curve with mean 57.51\% and standard deviation 16.47 percentage points.
}
    \label{fig:acceptance-rate-distribution}
\end{subfigure}
\hfill
\begin{subfigure}[t]{0.4\linewidth}
    \centering
    \includegraphics[width=0.7\linewidth]{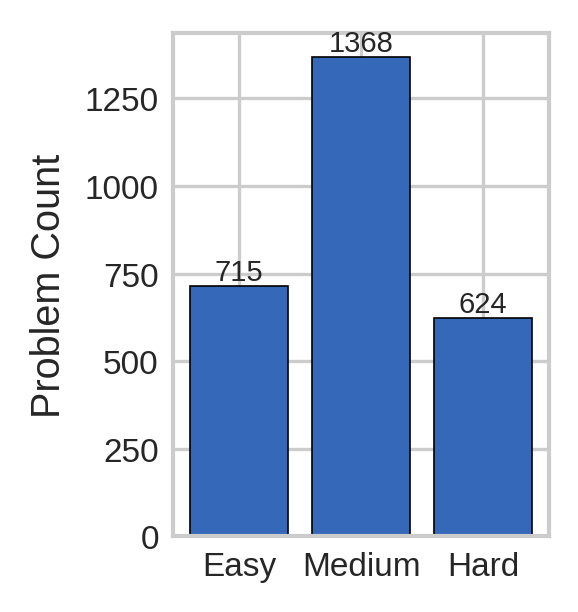}
    \caption{Problem difficulty distribution in the cleaned benchmark. Medium problems form the largest slice, followed by easy and hard problems.}
    \label{fig:corpus-difficulty}
\end{subfigure}

\caption{Benchmark problem characteristics: human acceptance-rate distribution and problem-topic distribution.}
\label{fig:benchmark-characteristics}
\end{figure*}

The benchmark is large and heterogeneous, but it is not uniform. Medium problems dominate the corpus, followed by easy and hard problems. Figure~\ref{fig:corpus-difficulty} summarizes the difficulty distribution of the cleaned archive. The distribution contains 715 easy problems, 1,368 medium problems, and 624 hard problems. This imbalance matters analytically because medium problems dominate aggregate scores, while the smaller hard subset tests whether model performance holds under higher algorithmic difficulty.

\begin{figure}[!tbp]
\centering

\includegraphics[width=\linewidth]{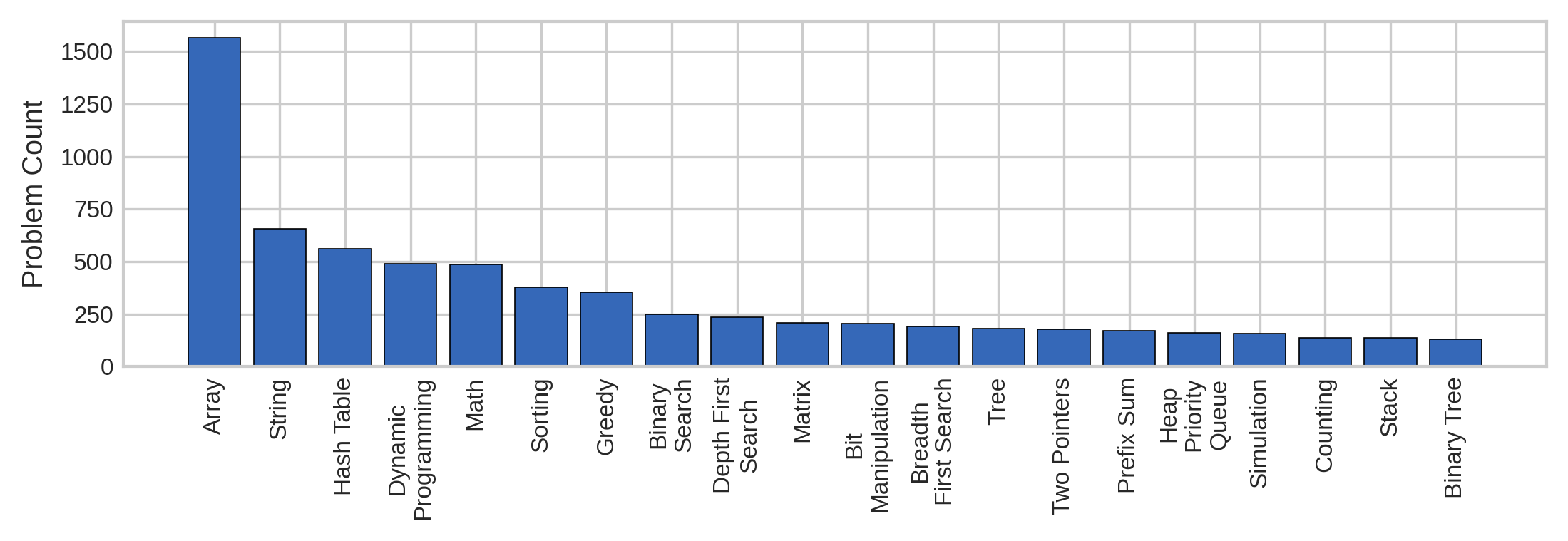}
    \caption{Problem-topic distribution. Array, string, hash-table, dynamic-programming, and math problems dominate the topic distribution.}
    \label{fig:topic-distribution}
\end{figure}

Language coverage is broad but not perfectly uniform. Some problems do not have support for all languages. JavaScript and TypeScript each cover 2,629 unique problems; C++ and Java each cover 2,598; C\# covers 2,597; Go covers 2,596; C covers 2,592; and PHP, Ruby, Scala, and Swift each cover 2,591. Python has lower retained coverage, with 1,728 unique problems. Human acceptance rates provide a second view of benchmark difficulty. Figure~\ref{fig:acceptance-rate-distribution} shows the distribution of human acceptance rates over the problem set. The fitted distribution has a mean of 57.51\% and a standard deviation of 16.47 percentage points, indicating substantial variation in how difficult problems are for human solvers. Most problems cluster in the mid-range, with many falling between roughly 40\% and 75\% acceptance, while fewer problems appear in the very low-acceptance or very high-acceptance tails. This distribution provides context for interpreting model correctness because model performance can be compared against a human-centered difficulty signal rather than only against aggregate pass rates. The topic distribution reflects the composition of the available LeetCode problem set used in this study. Figure~\ref{fig:topic-distribution} shows that arrays are the most common topic, followed by strings, hash tables, dynamic programming, and math. Consequently, aggregate results are shaped more by high-frequency topics than by lower-frequency graph, heap, or advanced data-structure topics. Arrays remain the dominant topic.

\subsection{Model Performance}
\label{sec:model-performance}

\begin{figure*}[!tbp]
\centering

\begin{subfigure}[t]{0.48\linewidth}
    \centering
    \includegraphics[width=\linewidth]{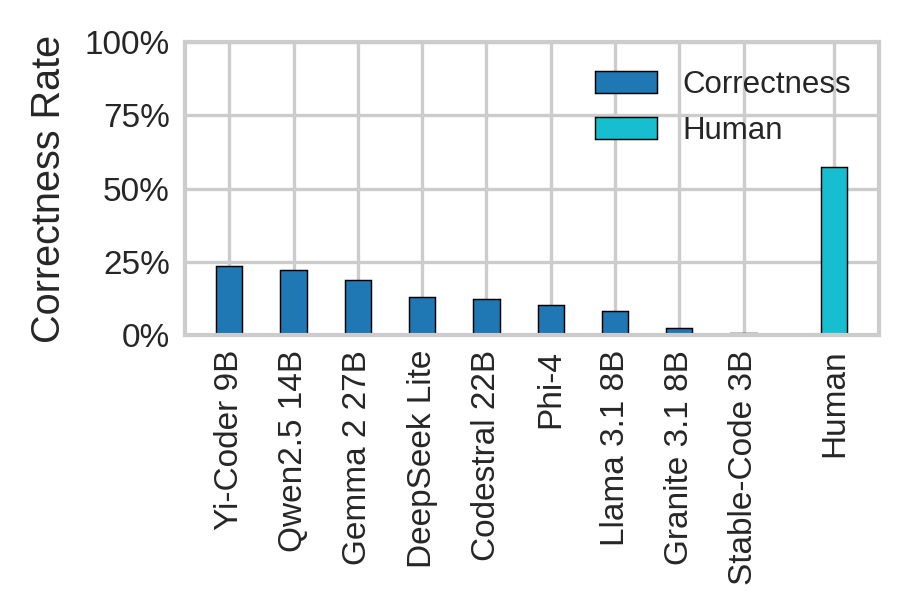}
    \caption{Overall functional performance. Mean correctness is shown for each model and compared with the human acceptance baseline.}
    \label{fig:performance-overall}
\end{subfigure}
\hfill
\begin{subfigure}[t]{0.48\linewidth}
    \centering
    \includegraphics[width=\linewidth]{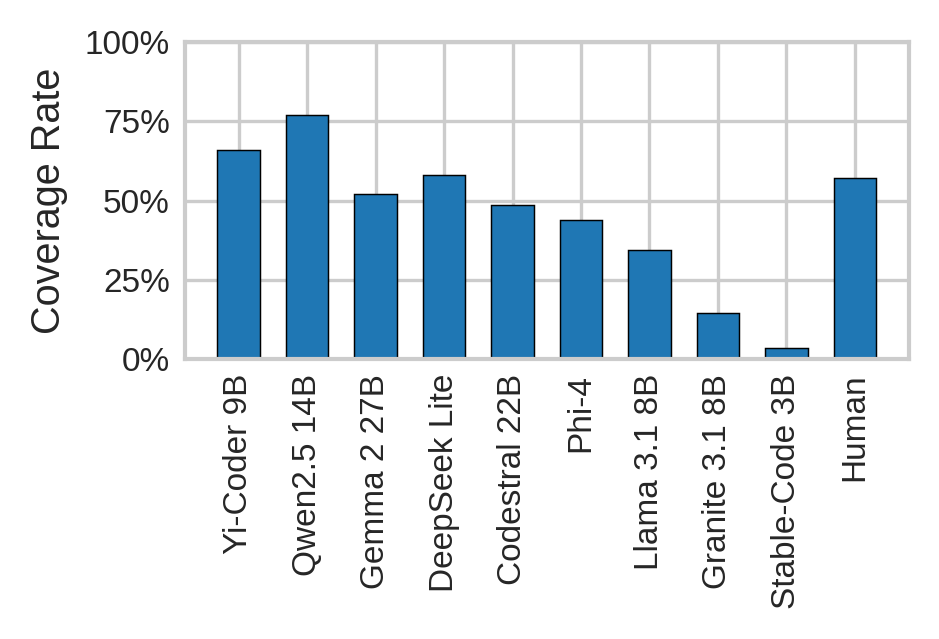}
    \caption{Accepted-problem coverage by model. Qwen covers the broadest set of problems despite not having the highest mean correctness.}
    \label{fig:accepted-problem-coverage}
\end{subfigure}

\caption{Comparison of overall functional performance and accepted-problem coverage across models.}
\label{fig:performance-and-coverage}
\end{figure*}

Figure~\ref{fig:performance-overall} compares global model correctness with the human acceptance baseline. Yi-Coder-9B-Chat has the strongest mean correctness, with Qwen2.5-Coder-14B-Instruct and Gemma-2-27B-IT forming the next tier. The absolute gap to the human acceptance reference remains large for every model, showing that open coding models still solve only a minority of the LeetCode problems under this single-pass setting. The figure is therefore best read as a high-level leaderboard.

Figure~\ref{fig:accepted-problem-coverage} reports accepted-problem coverage, which measures breadth rather than average correctness across all jobs. This view changes the ordering, where Qwen covers the largest share of accepted problems, indicating that a model can solve a broader set of distinct problems even when its average job-level correctness is slightly lower than the top mean performer. Coverage is useful because a downstream user may care less about repeated success on the same problem across languages and more about whether the model can find at least one accepted solution for many different problems.

\begin{figure*}[!tbp]
\centering
\includegraphics[width=0.7\linewidth]{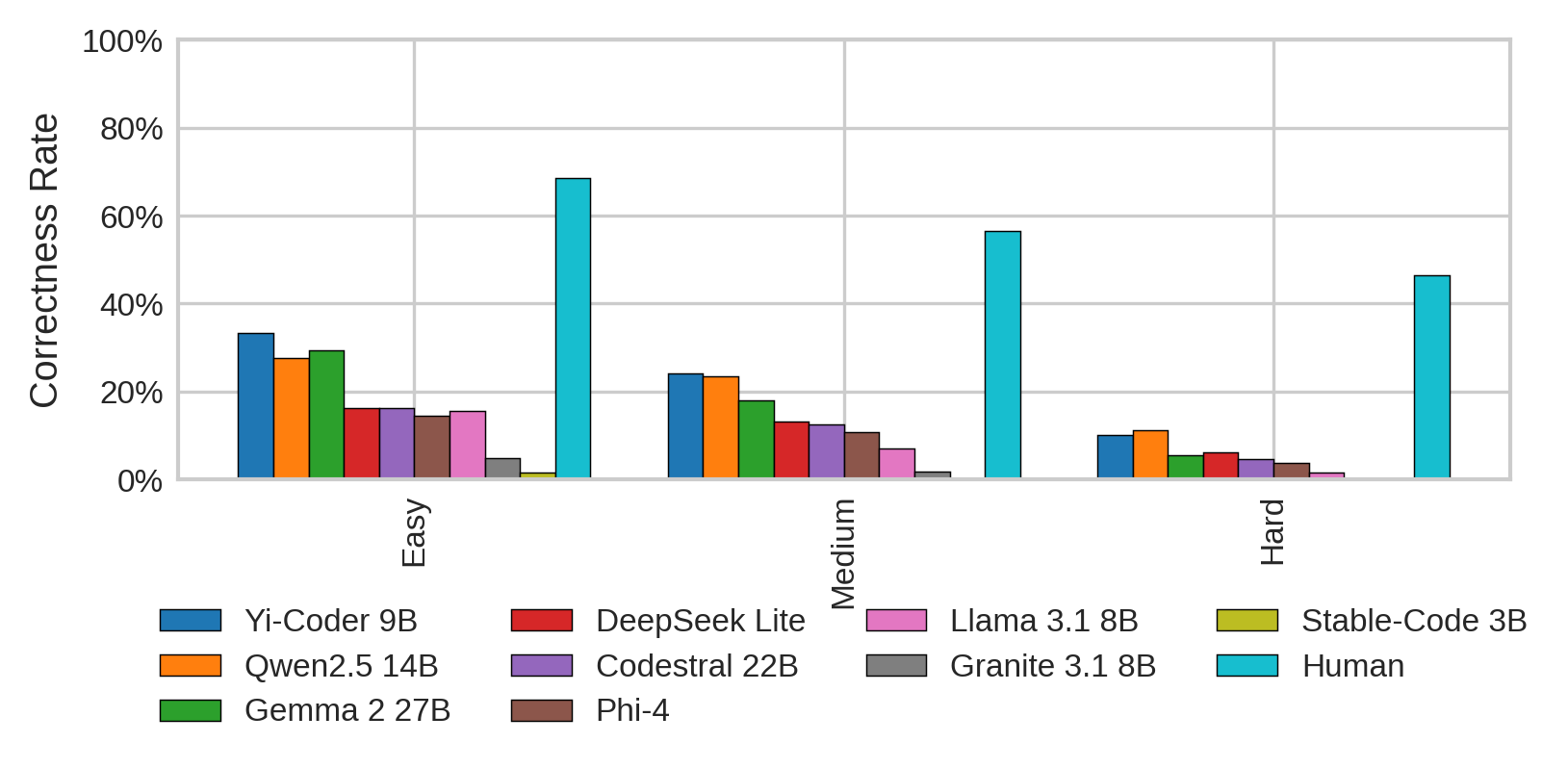}
\caption{Model correctness stratified by problem difficulty. Performance declines sharply from easy to hard problems, and the strongest model differs by slice.}
\label{fig:performance-difficulty}
\end{figure*}

Difficulty-conditioned performance shows that model rankings are slice-dependent. Figure~\ref{fig:performance-difficulty} shows Yi-Coder leading on easy and medium problems, while Qwen is strongest on hard problems. The drop from easy to hard is steep for every model, confirming that the benchmark is not saturated. This suggests that model comparisons should not be reduced to a single global score when the downstream workload emphasizes harder tasks.

\begin{figure*}[!tbp]
\centering
\includegraphics[width=0.8\linewidth]{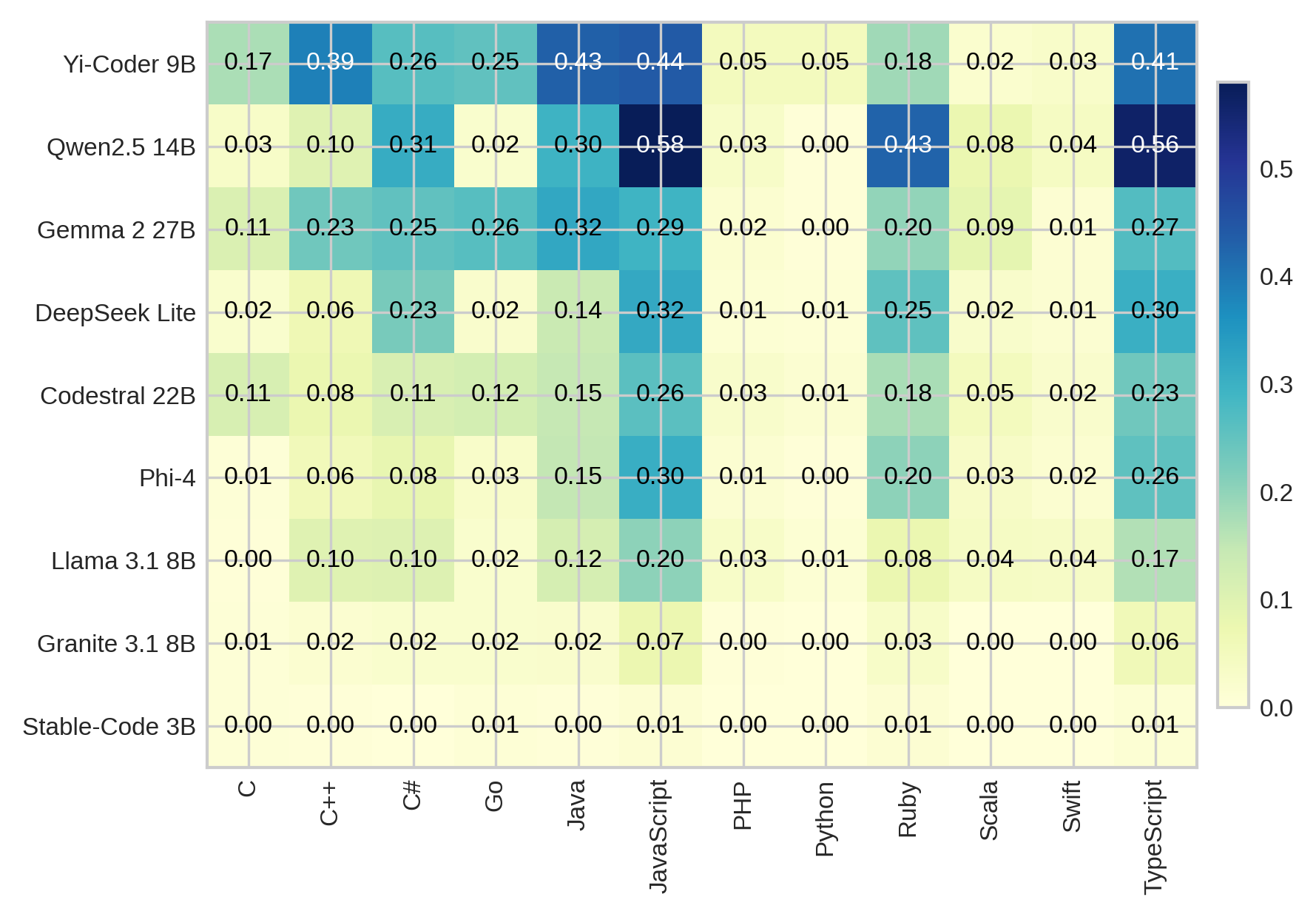}
\caption{Heatmap of model performance by programming language. The heatmap captures language-conditioned correctness, highlights cross-language variation, and makes model-specific strengths easier to compare than global averages.}
\label{fig:language-heatmap}
\end{figure*}

The language-conditioned results reveal another source of heterogeneity. Figure~\ref{fig:language-heatmap} reports correctness by target language and model, showing that model quality is not language-invariant. Some models are comparatively strong in conventional LeetCode languages such as C++, Java, and Python, while others are more competitive in languages such as C\#, Swift, Ruby, or TypeScript. The heatmap also makes the relative winners easier to compare across language--model pairs. Yi-Coder is strongest for several conventional compiled or scripting targets, Qwen is especially competitive on C\#, JavaScript, Ruby, Swift, and TypeScript, and Gemma is strongest on Go and Scala. Thus, model ranking depends substantially on the target programming language, reinforcing the central premise of this study that multilingual coding benchmarks can yield conclusions different from single-language evaluations.

Topic-conditioned performance shows similar non-uniformity. Table~\ref{tab:model-topic-performance} reports correctness across the 20 most common LeetCode topic tags. Yi-Coder is consistently strong across many high-frequency topics, while Qwen remains highly competitive and leads selected algorithmic categories. The table also shows that topic-level rankings do not preserve a single global ordering: for example, two-pointers, stack, binary-tree, and graph-search topics expose different relative strengths across models.

\begin{table*}[t]
\centering

\setlength{\tabcolsep}{4pt}
\caption{Correctness by topic for the 20 most common topics. Values are percentages.}
\label{tab:model-topic-performance}
\resizebox{0.9\textwidth}{!}{%
\begin{tabular}{lrrrrrrrrrr}
\toprule
Topic & Yi-Coder & Qwen2.5 & Gemma & DeepSeek & Codestral & Phi-4 & Llama3.1 & Granite3.1 & StableCode & Human \\
\midrule
array & 21.9 & 21.4 & 17.7 & 11.9 & 10.3 & 9.6 & 7.3 & 2.1 & 0.4 & 56.0 \\
string & 23.7 & 20.3 & 18.5 & 12.0 & 11.9 & 9.2 & 8.2 & 2.4 & 0.6 & 57.2 \\
hash-table & 23.0 & 22.8 & 16.1 & 12.7 & 11.8 & 10.1 & 6.5 & 2.2 & 0.4 & 56.9 \\
dp & 18.9 & 15.3 & 14.2 & 10.2 & 8.5 & 7.2 & 5.1 & 1.1 & 0.1 & 49.4 \\
math & 21.2 & 19.7 & 16.3 & 10.6 & 10.6 & 8.5 & 7.5 & 2.0 & 0.5 & 54.9 \\
sorting & 21.6 & 22.4 & 16.9 & 11.7 & 9.3 & 8.6 & 6.4 & 2.1 & 0.3 & 56.8 \\
greedy & 18.3 & 18.6 & 12.5 & 8.6 & 6.3 & 6.1 & 5.2 & 0.7 & 0.1 & 53.7 \\
binary-search & 21.1 & 20.3 & 18.6 & 12.7 & 12.3 & 9.9 & 7.4 & 2.6 & 0.4 & 47.8 \\
depth-first-search & 26.6 & 23.5 & 15.5 & 13.2 & 16.3 & 11.6 & 7.9 & 1.9 & 0.0 & 60.8 \\
matrix & 22.2 & 21.9 & 18.7 & 13.1 & 12.2 & 11.2 & 7.1 & 2.2 & 0.1 & 61.2 \\
bit-manipulation & 18.2 & 18.8 & 14.1 & 11.0 & 10.7 & 8.7 & 6.5 & 2.2 & 0.7 & 57.8 \\
breadth-first-search & 25.1 & 23.7 & 15.6 & 14.0 & 15.0 & 11.2 & 7.4 & 1.4 & 0.0 & 59.3 \\
tree & 24.9 & 22.0 & 12.9 & 11.2 & 15.2 & 10.4 & 8.4 & 2.0 & 0.1 & 63.4 \\
two-pointers & 31.6 & 26.4 & 27.5 & 17.0 & 18.5 & 13.8 & 13.6 & 4.1 & 1.3 & 57.4 \\
prefix-sum & 15.1 & 17.2 & 12.4 & 8.8 & 7.7 & 8.0 & 5.9 & 1.6 & 0.8 & 53.7 \\
heap-priority-queue & 14.7 & 15.3 & 9.8 & 9.3 & 6.2 & 5.5 & 2.6 & 1.1 & 0.1 & 54.0 \\
simulation & 23.3 & 23.1 & 19.4 & 10.2 & 10.0 & 10.3 & 8.3 & 2.4 & 0.5 & 65.9 \\
counting & 24.0 & 23.1 & 17.2 & 11.3 & 8.4 & 7.9 & 7.1 & 2.9 & 0.4 & 60.2 \\
stack & 28.9 & 22.3 & 18.5 & 14.0 & 15.2 & 12.1 & 9.9 & 1.5 & 0.3 & 58.2 \\
binary-tree & 30.1 & 25.7 & 16.1 & 13.2 & 18.7 & 13.2 & 11.3 & 2.6 & 0.2 & 66.9 \\
\bottomrule
\end{tabular}%
}
\end{table*}

\begin{figure*}[!tbp]
\centering
\includegraphics[width=0.6\linewidth]{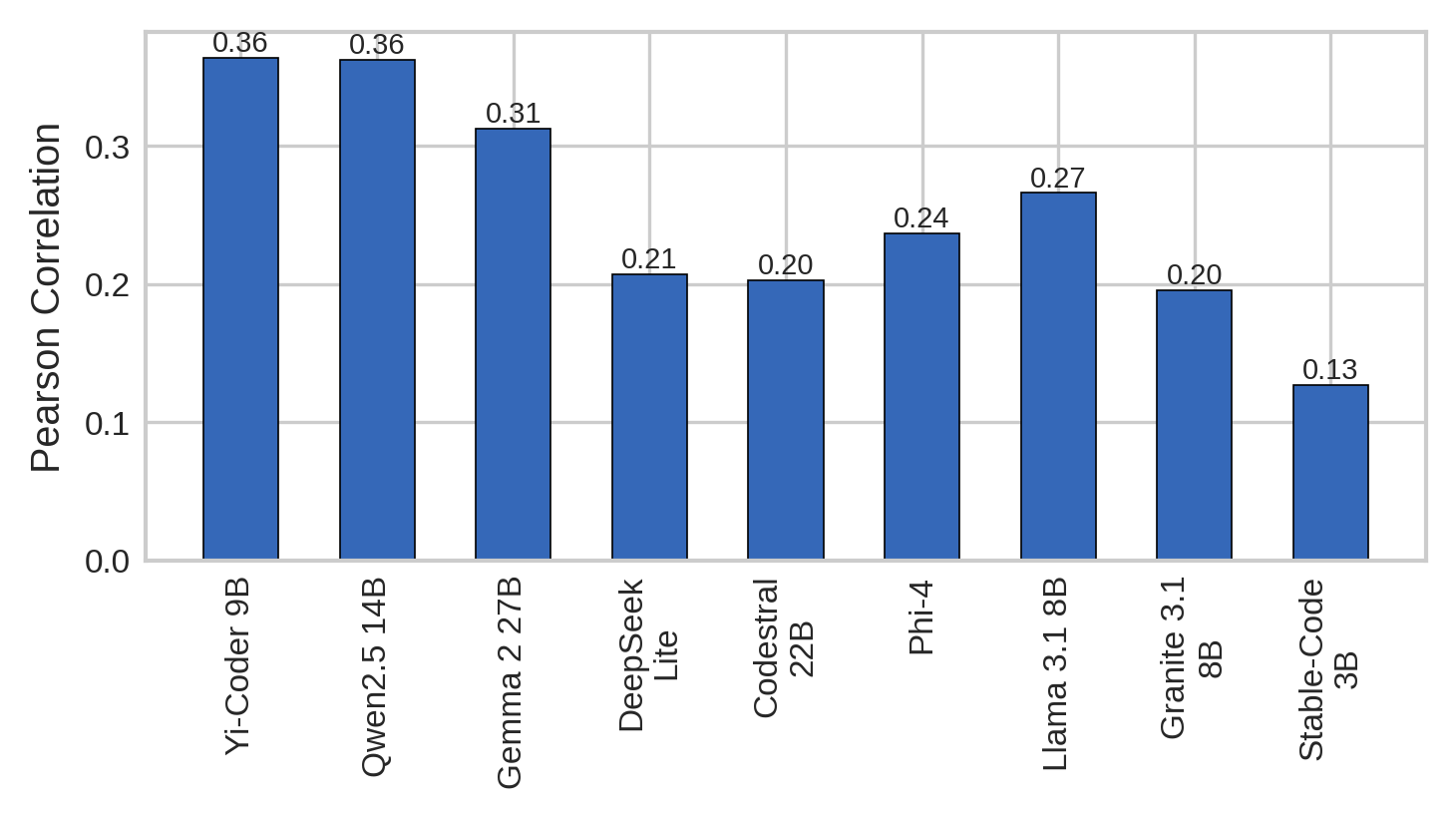}
\caption{Pearson correlation between model correctness and human acceptance rate. Stronger models correlate more with human acceptance, but the correlations remain moderate.}
\label{fig:acceptance-correlation}
\end{figure*}

\begin{figure*}[!tbp]
\centering
\includegraphics[width=0.6\linewidth]{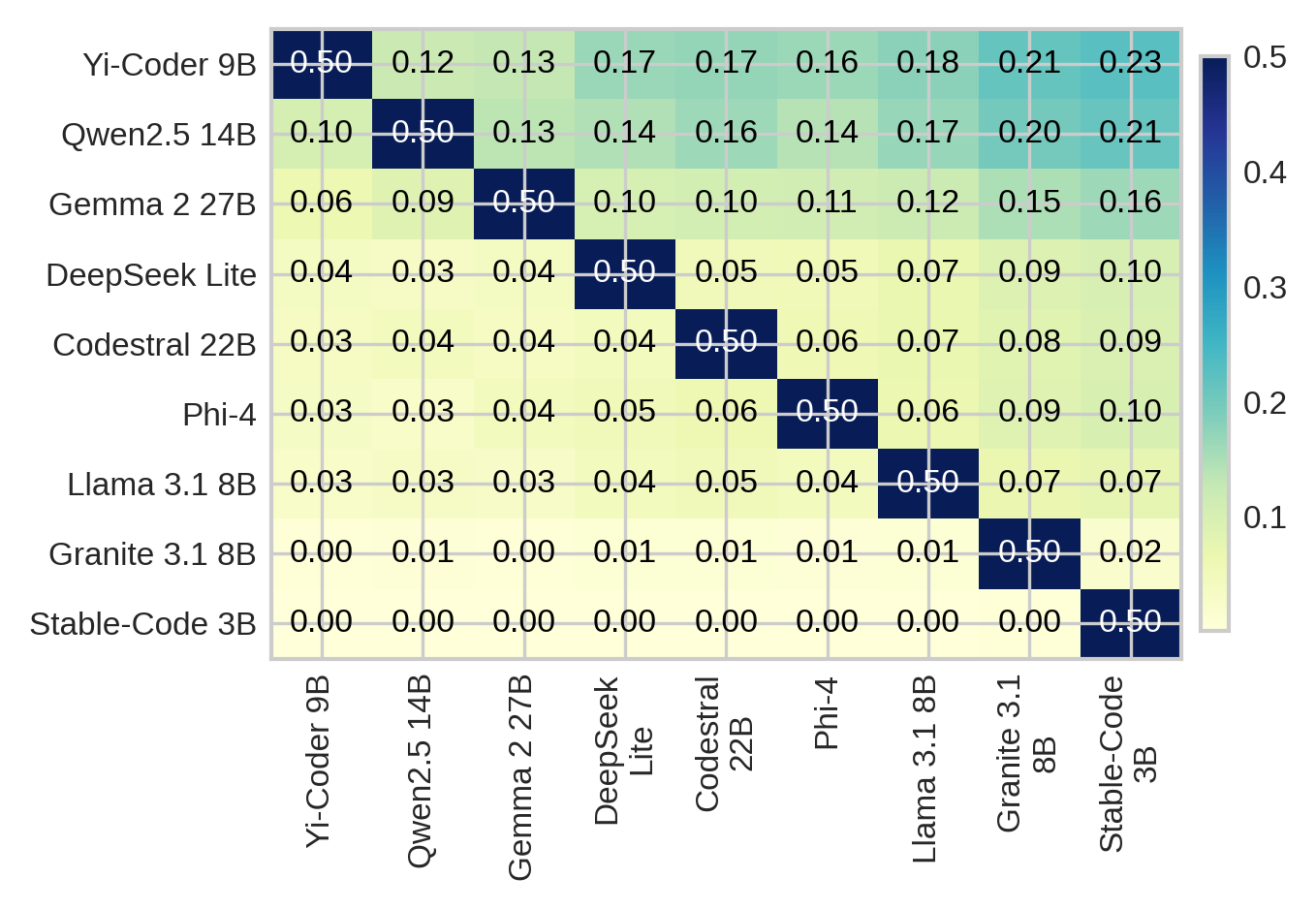}
\caption{Pairwise head-to-head win rates between models. Leading systems win more shared comparisons, while weaker models have low off-diagonal win rates.}
\label{fig:pairwise-head-to-head}
\end{figure*}

\begin{table*}[!tbp]
\centering
\small
\caption{Model summary. Correctness, hard-slice correctness, coverage, and lint pass are percentages. Quality is reported on a 0--1 scale.}
\label{tab:model-summary}
\resizebox{0.8\textwidth}{!}{%
\begin{tabular}{lrrrrrr}
\toprule
Model & Correctness & Hard & Coverage & Avg. Lang. Rank & Quality & All-Lang. Lint Pass \\
\midrule
Yi-Coder 9B & 23.6 & 10.1 & 66.0 & 2.33 & 0.519 & 85.7 \\
Qwen2.5 14B & 22.2 & 11.1 & 77.1 & 2.67 & 0.490 & 49.3 \\
Gemma 2 27B & 18.9 & 5.5 & 52.0 & 3.50 & 0.532 & 86.6 \\
DeepSeek Lite & 13.1 & 6.1 & 58.1 & 4.75 & 0.455 & 59.8 \\
Codestral 22B & 12.3 & 4.6 & 48.5 & 4.25 & 0.516 & 85.3 \\
Phi-4 & 10.3 & 3.7 & 43.8 & 5.33 & 0.461 & 45.0 \\
Llama 3.1 8B & 8.1 & 1.5 & 34.3 & 5.42 & 0.469 & 53.6 \\
Granite 3.1 8B & 2.6 & 0.3 & 14.6 & 7.92 & 0.476 & 79.9 \\
Stable-Code 3B & 0.7 & 0.0 & 3.6 & 8.83 & 0.588 & 83.2 \\
\bottomrule
\end{tabular}%
}
\end{table*}


Figure~\ref{fig:acceptance-correlation} shows that human acceptance rate provides a useful  signal of model success. The strongest associations are observed for Yi-Coder 9B and Qwen2.5 14B, both with Pearson correlations of 0.36, followed by Gemma 2 27B at 0.31. This indicates that these models are more likely to succeed on problems that are also easier for human LeetCode users. However, all correlations remain moderate, ranging from 0.13 to 0.36, which shows that human acceptance rate does not fully capture model difficulty. Therefore, acceptance-rate metadata is informative for interpreting benchmark behavior.

Pairwise comparisons give a direct view of relative model dominance on shared tasks. Figure~\ref{fig:pairwise-head-to-head} shows that Yi-Coder and Qwen dominate the head-to-head matrix, while weaker systems rarely beat the leading models. This reinforces the global ordering while still preserving the asymmetric nature of pairwise wins: a model can lose globally yet retain pockets of strength against another model on particular problem--language combinations.
\subsection{Failure Modes}
\label{sec:failure-analysis}

\begin{figure*}[!tbp]
\centering

\begin{subfigure}[t]{0.55\linewidth}
    \centering
    \includegraphics[width=\linewidth]{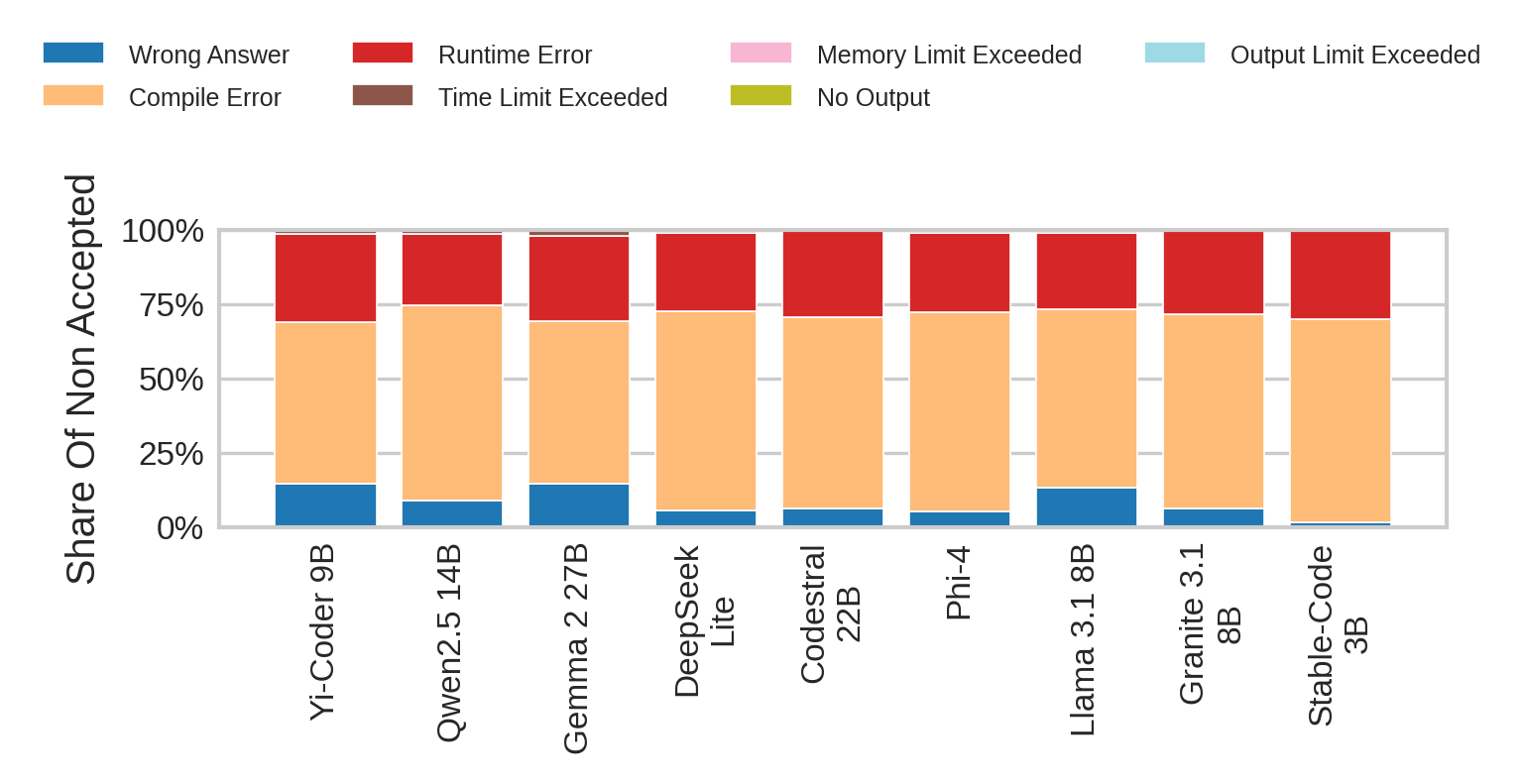}
    \caption{Non-accepted failure composition by model. Compile errors dominate the non-accepted subset, with smaller shares of wrong-answer, runtime, memory-limit, time-limit, no-output, and output-limit outcomes.}
    \label{fig:model-nonaccepted-breakdown}
\end{subfigure}
\hfill
\begin{subfigure}[t]{0.44\linewidth}
    \centering
    \includegraphics[width=\linewidth]{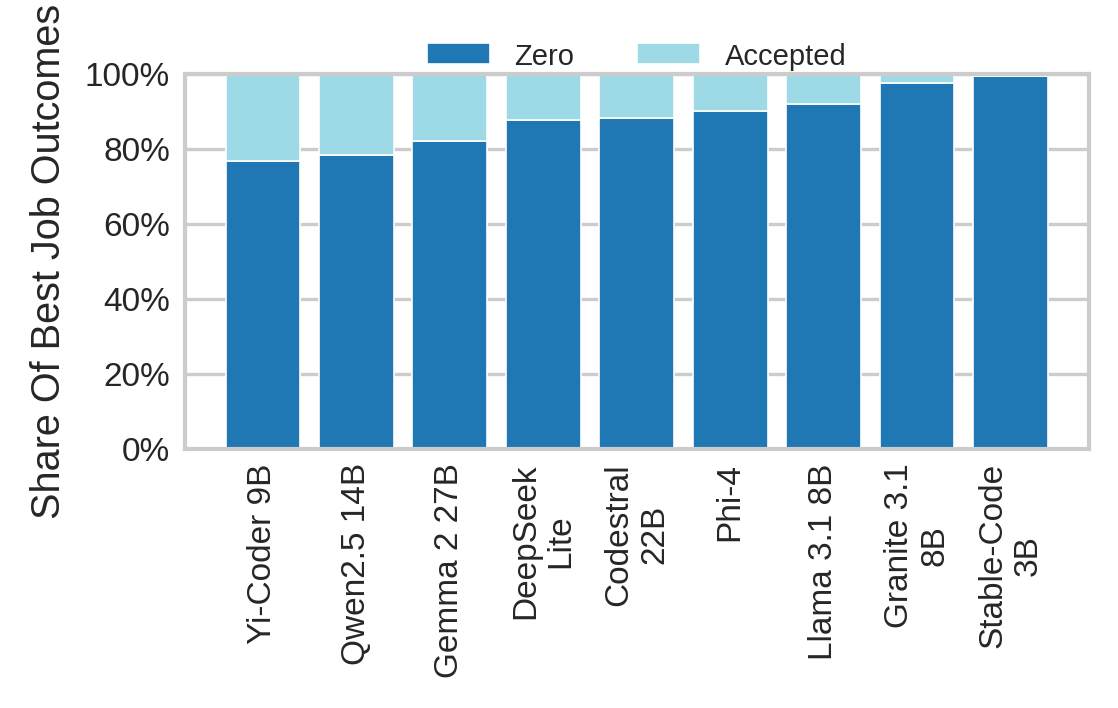}
    \caption{Partial-correctness distribution by model. The archive mainly separates zero-output wrong answers from accepted outputs, providing limited information about near-correct executions.}
    \label{fig:partial-correctness}
\end{subfigure}

\caption{Failure-mode composition and partial-correctness distribution across models.}
\label{fig:failure-and-partial-correctness}
\end{figure*}

In this part of the study, we analyze non-accepted submissions to understand where generated programs break down. Model-specific breakdowns further show that failure profiles differ across systems. Figure~\ref{fig:model-nonaccepted-breakdown} shows that compile errors are the leading non-accepted outcome for every model, while stronger models tend to reduce this share and expose relatively more runtime or wrong-answer cases. This shift is important because it separates failures that prevent execution from failures that occur after the program runs. From an operational perspective, improving coding models should therefore mean not only increasing accepted outputs, but also moving residual failures toward more diagnosable runtime or semantic errors.

The wrong-answer and partial-correctness results also reveal a limitation of the stored execution traces. Wrong-answer cases are concentrated in the zero-percent passed-testcase bin across models, with little evidence of intermediate testcase success. Figure~\ref{fig:partial-correctness} similarly shows that partial correctness is concentrated near the extremes rather than distributed across near-miss regions. Thus, the archive is useful for characterizing broad failure modes, but less effective for distinguishing almost-correct wrong answers from fully incorrect ones.


\subsection{Code Quality}
\label{sec:code-quality}

\begin{figure}[!tbp]
\centering
\includegraphics[width=\linewidth]{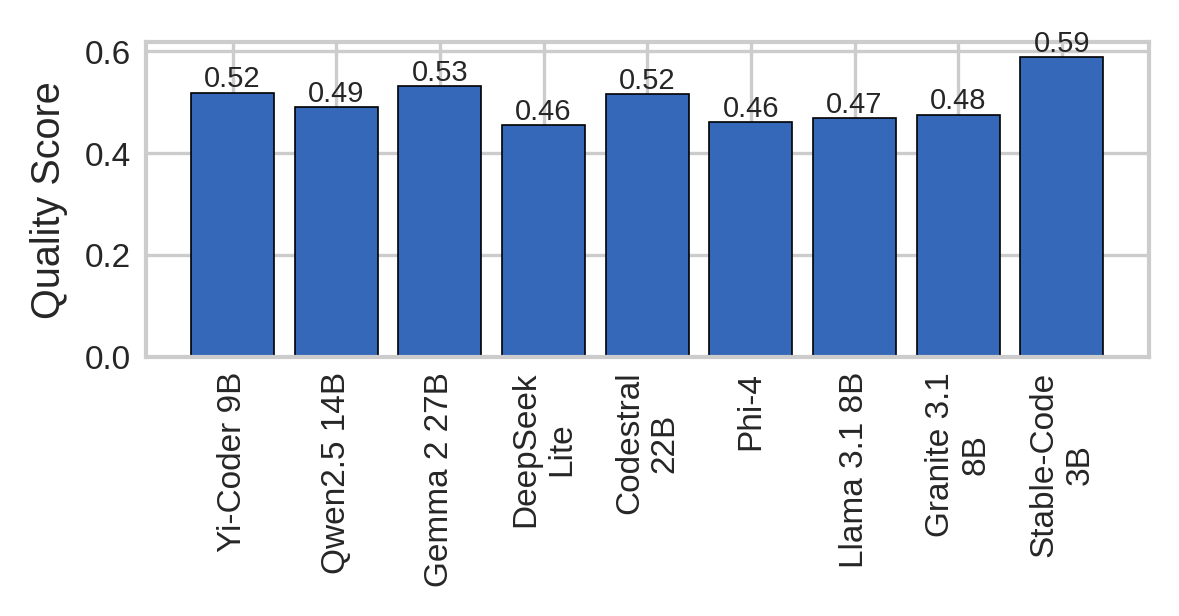}
\caption{Overall structural code-quality score by model.}
\label{fig:quality-structural-overall}
\end{figure}

In this part we analyze the code quality generated by the LLMs. Figure~\ref{fig:quality-structural-overall} reports the structural quality score, where Stable-Code-Instruct-3B obtains the strongest score despite the weakest functional performance in the benchmark. This shows that structural regularity does not imply executable correctness, since simple, low-complexity code may still fail the task.

\begin{figure*}[!tbp]
\centering

\begin{subfigure}[t]{0.48\linewidth}
    \centering
    \includegraphics[width=\linewidth]{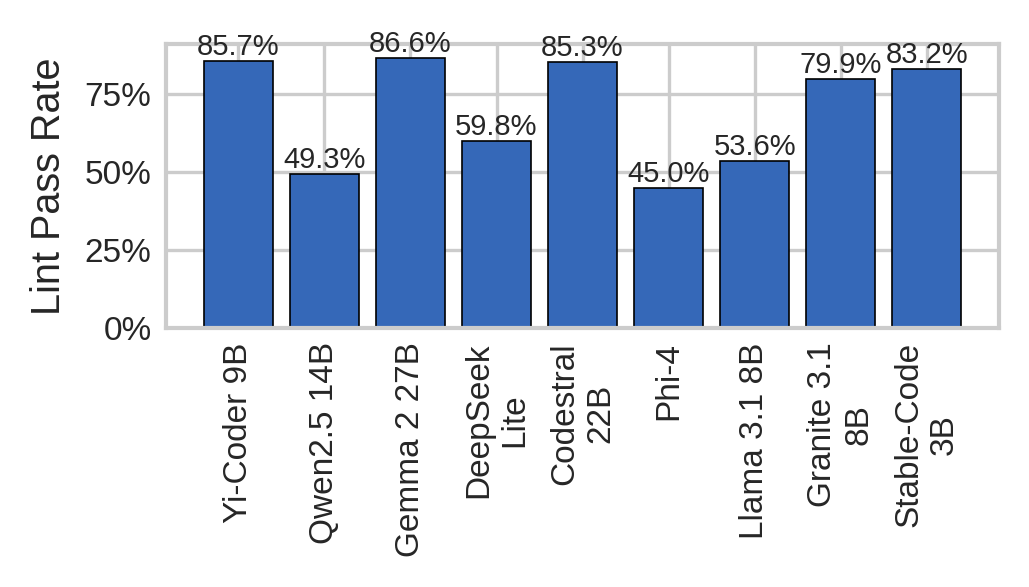}
    \caption{All-language lint pass rate across best-job code artifacts. Lint cleanliness and functional success produce different model rankings.}
    \label{fig:all-language-lint-pass}
\end{subfigure}
\hfill
\begin{subfigure}[t]{0.48\linewidth}
    \centering
    \includegraphics[width=\linewidth]{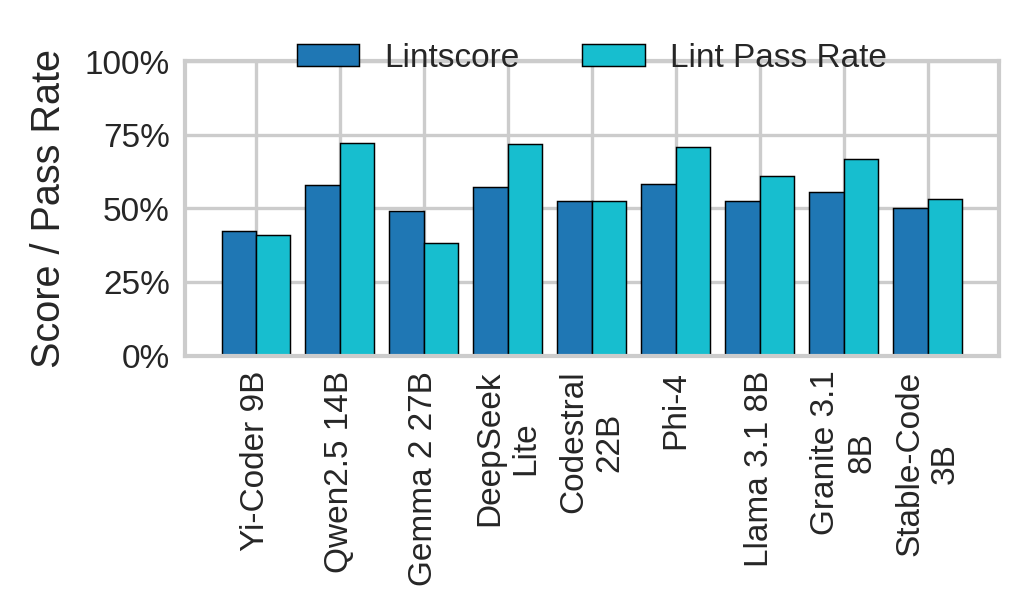}
    \caption{Accepted-code linter score and lint pass rate by model. This view estimates review burden among submissions that already achieved accepted status.}
    \label{fig:accepted-linter-overall}
\end{subfigure}

\caption{Linter-based evaluation across all best-job code artifacts and accepted-code submissions.}
\label{fig:linter-combined}
\end{figure*}

\begin{figure}[!tbp]
\centering
\includegraphics[width=0.9\linewidth]{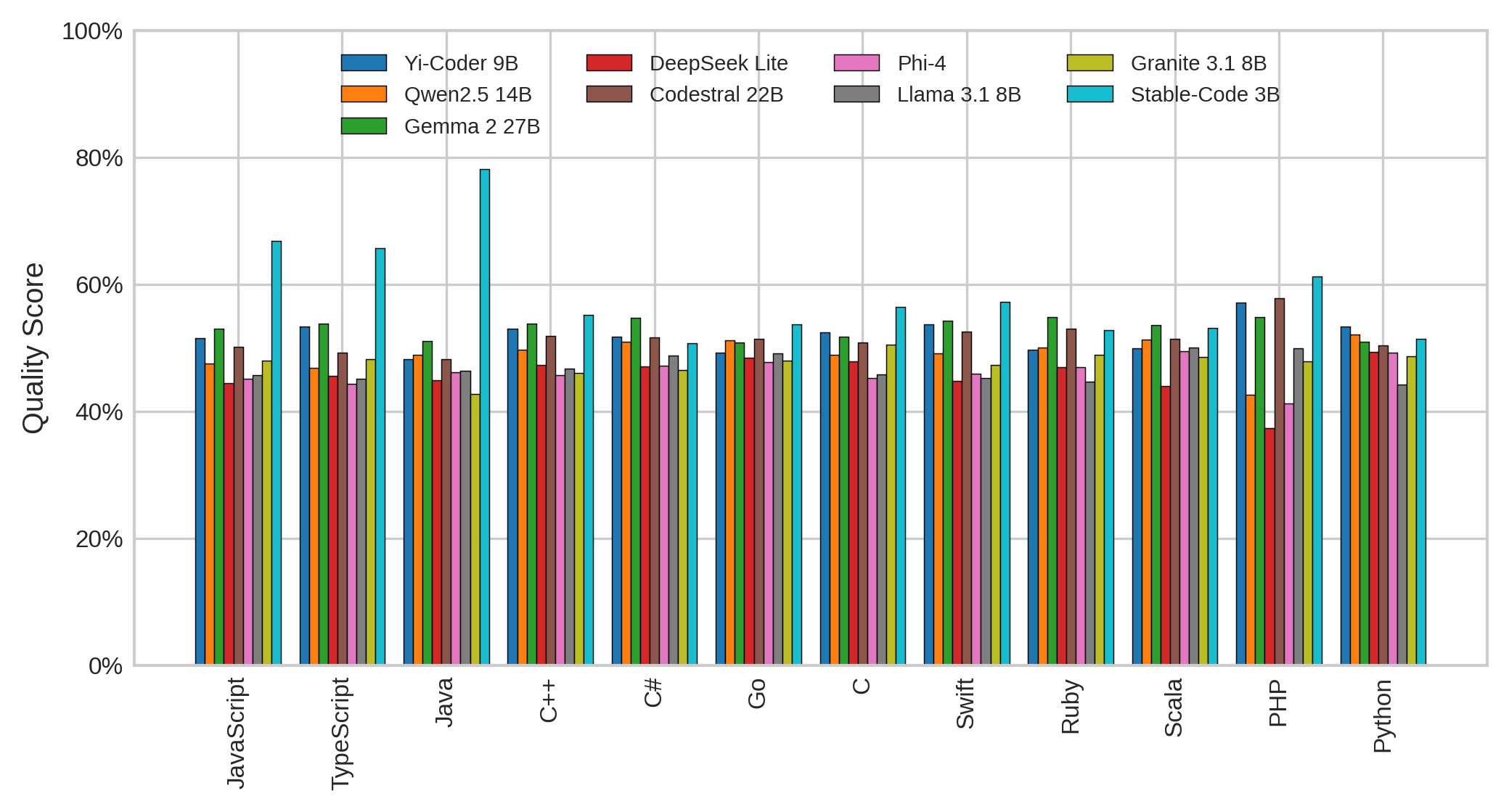}
\caption{Language-conditioned structural code quality by model, showing substantial variation in static quality across programming languages.
}
\label{fig:quality-by-language}
\end{figure}

Figure~\ref{fig:all-language-lint-pass} shows that Gemma, Yi-Coder, and Codestral achieve the strongest all-language lint pass rates, while Qwen passes lint rules less often despite strong execution performance. This confirms that functional correctness and static cleanliness measure different properties. Figure~\ref{fig:quality-by-language} further shows that static quality is language-dependent, since a model can produce cleaner code in one language than another. Accepted-code linter coverage is limited to JavaScript, TypeScript, Java, C++, Go, C, and Python, all with 100\% coverage, while Ruby, C\#, Scala, PHP, and Swift are uncovered due to usable linters. Thus, accepted-code linter conclusions apply only to the seven covered languages.

\begin{figure}[!tbp]
\centering
\includegraphics[width=0.8\linewidth]{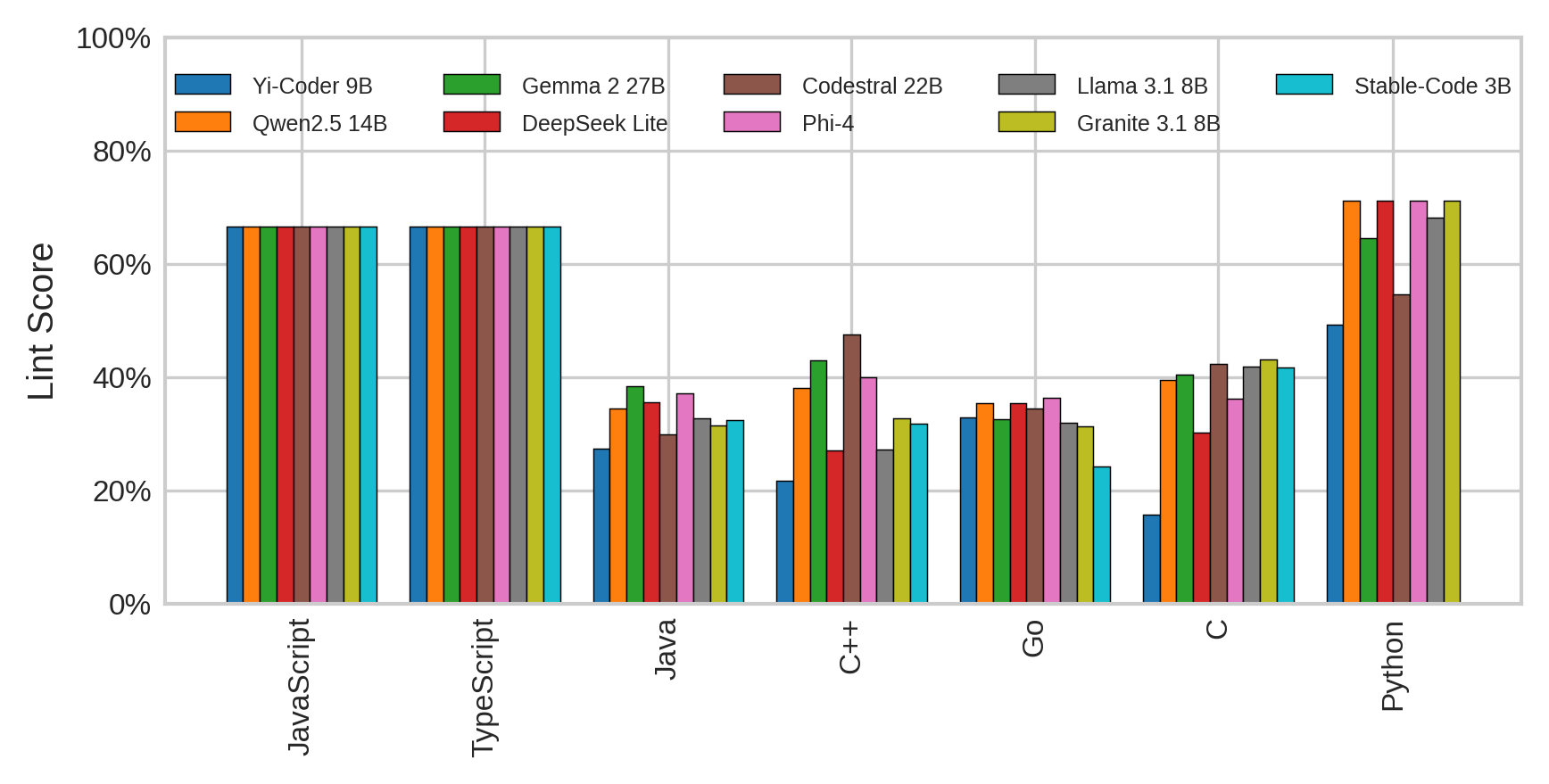}
\caption{Accepted-code linter score by language and model. Linter quality varies across both model family and programming language.}
\label{fig:accepted-linter-language}
\end{figure}

Figures~\ref{fig:accepted-linter-overall} and~\ref{fig:accepted-linter-language} analyze this covered subset. These results estimate review burden among code that already solves the task, showing that a functionally strong model may still require more cleanup. The language-specific view complements execution scores by showing where accepted solutions remain stylistically or structurally weaker.

All-language lint results show that the issue burden is dominated by warnings and conventions rather than hard errors or refactor findings. The aggregate category counts contain 141,376 warning findings, 45,537 convention findings, 2,280 error findings, and 402 refactor findings. Thus, the all-language static-analysis burden is driven primarily by warning-style and convention-style issues rather than refactor suggestions. Model-level category rates further show that models differ not only in lint-pass outcomes, but also in the types of static-analysis findings they accumulate. Warning rates are highest for Qwen2.5-Coder-14B-Instruct, Phi-4, and Meta-Llama-3.1-8B-Instruct, convention rates are more moderate, and refactor rates remain near zero across all models.

Table~\ref{tab:linter-issue-shares} summarizes the dominant convention and error subtypes previously reported as separate figures. Convention issues are concentrated in namespace and wildcard-import style patterns, with using-namespace-std accounting for roughly 48\% of convention findings, wildcard import about 21\%, force unwrap about 17\%, var about 6\%, global about 4\%, and global-variable/var-declaration patterns making up most of the remaining mass. Error findings are dominated by AST-grep failures at about 50\%, followed by throw-error patterns at about 15\%, fatal-error and throw-exception patterns at roughly 7\% each, panic at about 6\%, and raise/throw-exception variants in the lower single digits. These patterns are more actionable than a single lint score because they indicate what kind of cleanup a developer or repair system would need to perform.

\begin{table*}[!t]
\centering
\scriptsize
\caption{Dominant all-language linter issue shares. Shares are within each linter category and replace the separate convention/error share plots.}
\label{tab:linter-issue-shares}
\resizebox{0.6\textwidth}{!}{%
\begin{tabular}{llr}
\toprule
Category & Issue subtype & Share \\
\midrule
Convention & using-namespace-std & $\approx 48\%$ \\
Convention & wildcard import & $\approx 21\%$ \\
Convention & force unwrap & $\approx 17\%$ \\
Convention & var & $\approx 6\%$ \\
Convention & global & $\approx 4\%$ \\
Convention & global-variable / var-declaration & small remainder \\
\midrule
Error & AST-grep failure & $\approx 50\%$ \\
Error & throw-error & $\approx 15\%$ \\
Error & fatal-error & $\approx 7\%$ \\
Error & throw-exception & $\approx 7\%$ \\
Error & panic & $\approx 6\%$ \\
Error & raise / throw-exception variants & lower single digits \\
\bottomrule
\end{tabular}%
}
\end{table*}

Warnings are concentrated in output and logging patterns. Print-related findings form the largest warning subtype at about 25\% of warning issues, followed by system-output at about 12\%, println at about 10\%, console-log variants at roughly 9--10\% each, console-write-line and debug-print near 8\% each, var-dump around 4\%, and smaller print or debug-print variants below 2\%. Refactor findings are rare in absolute terms, with 402 total cases, and are dominated by boolean-comparison suggestions. The largest boolean-comparison subtype accounts for roughly 37\% of refactor findings, followed by two additional boolean-comparison subtypes near 22\% each, while the remaining variants appear only in small single-digit shares. These findings show that much of the all-language lint burden may not prevent execution, but it remains relevant for maintainability, output discipline, and production readiness.

The gap between execution behavior and static analysis is one of the main findings. A model can generate code that is short, regular, and structurally tidy while still failing to compile or solve the task. Conversely, a model that solves more problems may accumulate more lint findings because it produces more ambitious code or relies on conventions that work operationally but violate the shared rule set. Evaluation should therefore preserve both execution and static-quality evidence. The best model depends on the downstream goal, whether that goal is maximizing solved tasks, minimizing review burden, or balancing both.

\FloatBarrier

\section{Conclusion}
\label{sec:conclusion}

We presented a large-scale, multilingual, execution-grounded evaluation of openly accessible LLMs on LeetCode-derived programming tasks. By preserving prompts, raw responses, extracted code, official execution outcomes, and static-analysis signals, the benchmark supports analysis beyond aggregate pass rates. The results show that model comparison is multidimensional. Yi-Coder-9B-Chat achieves the strongest mean correctness and average language rank, Qwen2.5-Coder-14B-Instruct performs best on hard problems and accepted-problem coverage, and Gemma-2-27B-IT is strongest under the all-language lint pass-rate metric. Thus, the preferred model depends on whether the objective is average correctness, coverage, hard-task success, or reduced review burden.

The evaluation also exposes persistent reliability gaps. All models remain below the human acceptance reference, with compile errors dominating non-accepted best submissions. Overall, this study shows that evaluating code-generating LLMs requires more than identifying a single best model. Model performance depends on programming language, problem difficulty, topic composition, execution behavior, and code quality signals. By preserving the full generation-to-execution pipeline, this benchmark provides a reproducible basis for understanding not only when models succeed, but also how and why they fail. These findings highlight the need for execution-grounded, multilingual, and artifact-preserving evaluations as coding models continue to be used in increasingly diverse programming contexts.


\bibliographystyle{IEEEtran}
\bibliography{sn-bibliography}

\end{document}